\documentclass[preprint,5p,times,twocolumn,number]{elsarticle}

\usepackage{amssymb}
\usepackage{lipsum}
\usepackage[colorinlistoftodos]{todonotes}
\usepackage{color,soul}
\usepackage{amsmath}
\usepackage{svg}
\usepackage[stable]{footmisc}
\usepackage{float}
\usepackage{soul}
\usepackage{acronym} 

\DeclareMathOperator*{\argmin}{argmin} 

\journal{Applied Energy}

\newcommand{\cn}[1]{[citation needed]}

\begin{document}

\begin{frontmatter}


\author[vicomtech,polimi]{Julen Cestero\corref{l1}}
\ead{julen.cestero@polimi.it}
\cortext[l1]{Corresponding author}

\author[vicomtech]{Carmine Delle Femine}
\author[vicomtech]{Kenji S. Muro}
\author[vicomtech]{Marco Quartulli}
\author[polimi]{Marcello Restelli}
\affiliation[vicomtech]{organization={Vicomtech},
           addressline={}, 
           city={Donostia - San Sebastian},
           postcode={20009}, 
           state={Gipuzkoa},
           country={Spain}}
\affiliation[polimi]{organization={Politecnico di Milano},
           addressline={}, 
           city={Milano},
           postcode={20156}, 
           state={Lombardy},
           country={Italy}}

\title{Optimizing Energy Management of Smart Grid using Reinforcement Learning aided by Surrogate models built using Physics-informed Neural Networks}


\begin{abstract}
Optimizing the energy management within a smart grids scenario presents significant challenges, primarily due to the complexity of real-world systems and the intricate interactions among various components. 
\ac{RL} is gaining prominence as a solution for addressing the challenges of  \ac{OPF} in smart grids. However, \ac{RL} needs to iterate compulsively throughout a given environment to obtain the optimal policy. This means obtaining samples from a, most likely, costly simulator, which can lead to a sample efficiency problem. In this work, we address this problem by  
substituting costly smart grid simulators with surrogate models built using \ac{PINN}s, optimizing the RL policy training process by arriving to convergent results in a fraction of the time employed by the original environment.
Specifically, we tested the performance of our \ac{PINN} surrogate against other state-of-the-art data-driven surrogates and found that the understanding of the underlying physical nature of the problem makes the \ac{PINN} surrogate the only method that we studied capable of learning a good RL policy, in addition to not having to use samples from the real simulator.
Our work shows that, by employing \ac{PINN} surrogates, we can improve training speed by 50\%, comparing to training the RL policy by not using any surrogate model, enabling us to achieve results with score on par with the original simulator more rapidly.

\end{abstract}




\end{frontmatter}
\begin{acronym}[MPC] 
\acro{RL}{Reinforcement Learning}
\acro{EA}{Expert agent}
\acro{PINN}{Physics-Informed Neural Networks}
\acro{ANN}{Artificial Neural Network}
\acro{OPF}{Optimal Power Flow}
\acro{ESS}{Energy Storage Systems}
\acro{SoC}{State of Change}
\acro{MAE}{Mean Absolute Error}
\end{acronym}



\section{Introduction}\label{introduction}



Smart grids are a pivotal concept driving the current modernization of electrical networks, addressing the urgent need to reduce greenhouse gas emissions, enhance energy efficiency, and improve grid stability through demand response mechanisms. The European Union aims to achieve 43\% renewable energy generation by 2030 \cite{eugreendeal}, and in 2021, the renewable energy share rose to $32.1$\% \cite{euenergyshare}.

Modern societies require advanced grids capable of predicting and mitigating the uncertainties associated with renewable energy sources. These grids must leverage energy storage systems and demand response in an interoperable and manageable manner while ensuring security of supply and cost-effectiveness \cite{whataresgs}. In the European Union, over 13 countries have achieved smart meter installation rates exceeding 81\%, alongside the steady development of demand response and flexibility markets, as well as local energy communities \cite{lfmsuccess}.

On the one hand, the active monitoring and digitization capabilities provided by smart meters offer new opportunities for advanced control strategies based on demand response. These strategies adapt demand profiles by reducing consumption during peak hours and increasing it during off-peak periods \cite{drinsg}. On the other hand, simulating smart grids with distributed energy sources and energy storage systems has become increasingly complex, with rising computational demands \cite{sgrisks}.


Smart grid simulators enable the application of Machine Learning (ML) techniques and data-driven methods to explore optimal grid management strategies quantitatively. While traditional numerical optimization methods have been used for a long time for these purposes, integrating them with ML models, such as neural networks and decision trees, has proven effective for the predictive management of complex energy systems, including lithium batteries \cite{nnex1, nnex3} and large power grids \cite{rahmanMachineLearningAidedSecurity2020, rahmanLearningaugmentedApproachAC2021}.

\ac{RL} has demonstrated significant potential in developing efficient energy control systems with energy storage, leveraging its reward-focused approach \cite{rl4milk, rl4battinsg} and has ample capabilities in calculating optimal management strategies for power flow in smart grids. Furthermore, RL has been employed \cite{giraldo2023reinforcement} to model and design policies for demand response, using  Artificial Neural Networks (ANNs) to create accurate predictive models of demand response scenarios and by incorporating techniques like Kriging and Active Learning \cite{da2023surrogate}. 
However, RL needs constant interaction with the simulated system to converge to a functional policy. Applying RL to a real smart grid could lead to unproductive or risky states, even to the collapse of the grid itself during the policy training process. Simultaneously, simulators or simulated environments are crucial to addressing this issue.

In our work, we have employed the Gym-ANM framework, which supports the creation of detailed smart grid environments by incorporating physical constraints and diverse device types, including distributed generation and energy storage systems. This framework facilitates realistic power grid simulations and the training of optimal RL policies for efficient grid management \cite{anm6paper}.

Although simulators like these are essential for planning and managing smart grids, their simulation time and computational demands increase significantly with the complexity of the grid and its components, driven by the dimensional growth of variable matrices \cite{sgdimension}. 
To tackle these challenges, we propose using a Physics-Informed Neural Network (\ac{PINN}) surrogate as a replacement for the original smart grid environment. This surrogate model allows us to train a \ac{RL} policy without relying on the original environment, ultimately achieving a policy that converges to the performance score of the original system at a faster pace. Our findings highlight the critical importance of incorporating the physical nature of complex systems into modeling. This is demonstrated through a comparison with other state-of-the-art surrogate modeling methods that rely solely on data-driven approaches and do not leverage the underlying physical properties of the environment.

Our work shows that by employing surrogate \ac{PINN}s, we can accelerate the training process by 50\%, compared to training a \ac{RL} agent in the original environment without surrogation, enabling us to achieve the original environment's results more rapidly. This advancement has significant implications for the future development of Smart Grid management strategies and the integration of renewable energy sources into existing infrastructures.


\section{Optimal Power Flow and Control Systems in Smart Grids}
    Power system operators have been relying for a long time on \ac{OPF} techniques to calculate the energy generation system's most cost-effective dispatch to supply the energy demand of the grid in a stable manner, accounting for the technical constraints that its components require \cite{revml4opf}. Traditionally, \ac{OPF} problems were solved with numerical methods, mainly Newton-Raphson and other mathematical solutions based on recursive programming, Lagrange multipliers, or linearization, among others \cite{lagrangepoint, 6400272, revopfinsg}. However, with the change in the paradigm of the grid by the introduction of new renewable energy sources, distributed energy sources, and \ac{ESS}s, as well as schemes such as demand response and the overall shift towards a smart grid, new challenges have been created to solve \ac{OPF} problems mathematically due to new complex constraints and new dependencies in stochastic variables such as the weather or batteries' state of charge \cite{mlappsforopfrev, revopfinsg}.

    Some grid operators adopted simplifications of the \ac{OPF} problem from an AC-\ac{OPF} scheme into a DC-OPC scheme, which reduced computational expenses but returned suboptimal solutions that could pose safety threats or even generate unrealistic solutions \cite{revml4opf}. Data-driven machine learning methods, on the other hand, have large potential in addressing the computational requirements while maintaining stability by changing the strategy of online optimization to offline training by means of extensive historical or simulated data \cite{revml4opf}.

    Optimizing power flow and managing smart grids are computationally intensive tasks due to the complexity of detailed physical simulations. Surrogate models are increasingly used to approximate these complex systems, offering significant reductions in computation time \cite{mohammadi2024surrogate}. 
    However, the computational efficiency of surrogate models often involves accuracy trade-offs. Training surrogate models requires significant computational effort, especially when using high-fidelity simulations or physical constraints. This upfront cost can be justified by savings in inference time. Once trained, surrogate models provide rapid approximations, which is ideal for frequent recalculations or when direct simulation is too costly.
    
    Data-driven surrogates, such as decision trees, random forests, XGBoost, and deep neural networks, may perform well within the training data distribution but struggle with out-of-distribution scenarios or underrepresented regions.
    \ac{PINN}s offer an innovative approach that addresses many limitations of conventional surrogates \cite{raissi2019physics}. Unlike traditional data-driven models, \ac{PINN}s integrate physical equations, such as ordinary differential equations or partial differential equations, directly into the training loss function to ensure that the model's output remains consistent with fundamental physical laws, leading to improved robustness in poorly sampled regions of the state space compared to purely data-driven models. This makes them particularly suitable for modeling systems with high complexity and variability, such as energy networks with dynamic load profiles and renewable energy generation. For example, they have been used to model lithium batteries \cite{battPINN1, battPINN2} and its components such as electrodes \cite{battPINN3} and other \ac{ESS}s such as hydrogen electrifiers \cite{PINNs4h2}.

    Regarding smart grids, in \cite{PINNs4opf2} and \cite{PINNforopf}, \ac{PINN}s were developed, including active and reactive power balance equations to solve AC-\ac{OPF} problems, and in all study cases shown, \ac{PINN}s were more accurate than other conventional \ac{ANN} schemes.


    Furthermore, as \ac{OPF} is a problem distributed in an electrical grid topology,  Graph Neural Networks (GNNs) have also been used as surrogates. In \cite{unspgnn4opf}, a GNN trained in an unsupervised manner was compared with standard solvers to calculate power flows, and \cite{falconerLeveragingPowerGrid2023} compared a fully connected \ac{ANN}, a convolutional neural network and a GNN to predict the bus voltages and generator dispatch with varying load profiles, which showed that GNNs were much more capable of predicting accurately when the topology of the grid changed, which is frequent in large transmission grids.


    Other strategy to solve the \ac{OPF} problem is \ac{RL}, which is the strategy we followed in this work. The \ac{RL} approach focuses on optimizing the policy iteratively interacting with an environment. This environment, in this case a smart grid, returns a signal that shows the performance of the actions of the agent, and the algorithm aims to maximize this signal. Although many authors propose \ac{RL} as a solution for this problem \cite{drlex1, drlex2}, \ac{RL} needs to weigh the performance of the policy in several states of the environment and in complex environments, such as environments related to smart grids, this means that several samples are required from the environment, causing a loss in the training performance due to the curse of dimensionality \cite{barto2003recent}.

To address the fact that RL sample inefficiency compounded with the cost of multi-physics simulation limits its applicability, architectures employing ML-based surrogates of simulated environments have been proposed \cite{pinto2021data} \cite{kaewdornhan2022electric}.
We show that, in settings such as the optimal management of realistically complex energy networks, the accumulation of small (e.g., extrapolation) inaccuracies through episode histories limits the applicability of purely data-driven approaches. Hybrid models merging empirical learning with first principles are a conceivable remedy for this problem. In the present paper, we start to build a solid foundation for this approach by showing that \ac{PINN} models learned purely from first principles allow building RL systems whose learning performance significantly improves on the one obtainable from both expensive simulators and from inaccurate ML models.

The approach described in this paper continues the work of \cite{cestero2024building}. The authors contribute by analyzing methods for building surrogate models aimed at RL environments. Although they managed to build these surrogates accurately, they state a clear limitation of their work when using these surrogate models in an RL training process. This limitation is that the data-driven surrogate models --- trained using either actual transitions of the environment or generative transitions in the state space --- are not able to understand the underlying physical nature of the original environments, making these environments unreliable for obtaining the optimal policy via RL. In this work, we address this problem by using a novel method for building ANNs: PINNs. This approach is applied to a smart grid environment \cite{anm6paper}, and the results of this paper are used to contrast ours.

\section{Methods}

Our approach for solving the \ac{OPF} problem is using \ac{RL} for acquiring a working policy able to act in a smart grid well enough to minimize the penalties and the energy loss of the grid. For that, we use a smart grid simulator as a \ac{RL} environment: the ANM6-Easy environment \cite{anm6paper}. 

Although this environment is very precise with the fluctuations of the grid, costly mathematical iterations are needed to solve the equations that are contained in the environment. This makes this environment unfeasible for larger architectures, limiting the scalability of this solution via RL for solving the \ac{OPF} problem of a given smart grid. To solve this issue, we propose to use surrogate models as approximations to the environment. However, as described by \cite{cestero2024building}, using data-driven methodologies for building surrogates, although they seem to be robust for representing state transitions, showing good metrics in terms of R$^2$, sometimes this is not enough to represent precisely the physical nature of the state transitions. For that, we propose the use of \ac{PINN}s as a surrogate model, and we compare this method with other data-driven methodologies.

\begin{figure}
    \centering
    \includegraphics[width=\linewidth]{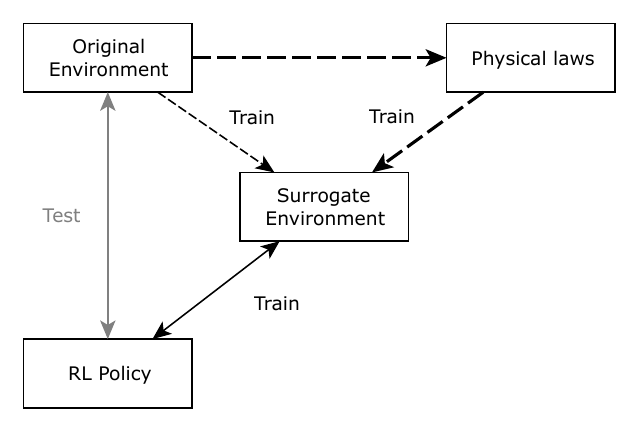}
    \caption{Architecture of our approach. There are two different methods for training the surrogate environment, used for either the \ac{PINN} method or the data-driven methods. The bold dashed lines show the \ac{PINN} training process, while the thin dashed line represents the data-driven process instead. Subsequently, the surrogate environment is used to train the RL policy, and after each policy update, the policy is tested against the original environment to get an accurate representation of the improvement of the policy.}
    \label{fig:architecture}
\end{figure}

The architecture of our solution is shown in Figure \ref{fig:architecture}. From the original environment, we study two different methods for training the surrogate environment: data-driven and \ac{PINN}-based. For the \ac{PINN} method, we obtain the physical laws from the original environments, and \textbf{without any use of the original environment}, we train a surrogate environment using the strategies described in \cite{femine2024kktinformed}. On the other hand, we use several state-of-the-art methods for building data-driven models, such as XGBoost, decision trees, etc. All the data-driven methods use the original environment to acquire a training dataset. Following the methods described by \cite{cestero2024building}, we obtain two different kinds of datasets: a so-called generative dataset, which basically acquires transitions from the environment independently from each other by calculating the transitions of a random sample of the state space of the environment; and an agent-based dataset, in which we used a random agent to acquire realistic trajectories from the environment following the policy of the agent. Using these two datasets separately, we train all the data-driven models. 

After training the surrogate environments using every model, we train a \ac{RL} policy using the PPO algorithm \cite{schulman2017proximal}. During the training, we test the policy after each update, running an episode on the original environment using the in-training policy. Afterward, the episode score is saved and used to study the evolution of the policies trained in each surrogate method.

    \subsection{Reinforcement Learning training characterization for surrogate models}

In this work, we use a surrogate model as the environment for the RL algorithms. To achieve this, we develop an algorithmic framework that connects the predictive model and the RL algorithms. This framework is organized following the guidelines of \texttt{gym} interfaces \cite{1606.01540}, which is required to interact with the PPO RL training algorithm \cite{schulman2017proximal} from the library Stable-baselines3 \cite{stable-baselines3} with default hyperparameter values. The transformations implemented within the framework are described as follows:

\begin{itemize}
    \item \textbf{Initial state selection}: the initial state is randomly sampled from the state space of the original environment. Each time the environment receives a reset signal, the initial state is determined as    $ \mathbf{s}_0 \sim \mathrm{Uniform}(\mathcal{S})$,
    where \( \mathcal{S} \) represents the entire set of possible states in the original environment.
    
    \item \textbf{State transition calculations}: the state transitions are computed using the surrogate model. The framework internally processes the current state and the action selected by the RL agent into a single input vector, formatted similarly to the input data structure of a predictive model as seen in \eqref{eq:surrogate}. The surrogate model predicts, then, the next state and the associated reward, following the function $    f: (\mathbf{s}_t, \mathbf{a}_t) \rightarrow (\mathbf{s}_{t+1}, r_t)$, 
    where \( f \) denotes the surrogate model
    and \( \mathbf{a}_t \in \mathcal{A}, \, \mathbf{a}_t \sim \pi(\mathbf{s}_t) \) an action selected from the action set by the policy \( \pi \).

    \item \textbf{Terminal state determination}: determining whether the environment has reached a terminal state, represented by the `done' flag  $d: d \sim \{0, 1\}$, cannot be performed directly as described in Section \ref{sec:terminal_state}, since the surrogate model is unable to discern a solvable and unsolvable set of equations \eqref{eq:active-power-balance} - \eqref{eq:reactive-power-balance}, always producing an output. For simplicity, a binary classifier based on gradient boosting \cite{Chen_2016} is used to identify terminal states with an accuracy of approximately 99\%. Details of the classifier training process are provided in \ref{sec:app_terminal_classificator}.
\end{itemize}

\subsection{Parallelization of the Surrogate Environment}

Using ANN-based surrogates enables us to exploit their natural parallelism rooted in linear algebra. In contrast to this, the original environment uses an internally iterative algorithm such as Newton-Raphson, which is inherently sequential and thus restricts parallel execution.

Leveraging this capability of the surrogate, we can significantly enhance the training speed of RL policies by enabling the parallelization of the environment. Parallelizing RL environments is arguably one of the most effective ways to optimize RL training, as it allows for the collection of environment samples at a much faster rate. This is achieved by running multiple episodes concurrently, without interference among them. 
Unfortunately, certain environments do not support parallelization. Such is the case for our target environment. However, creating an ANN-based surrogate of the environment fixes this problem.
This enhancement is achieved through a custom wrapper specifically designed to support the parallelization of surrogate environments. This wrapper works in conjunction with the framework defined in the previous section for transforming predictive models into surrogate environments.
The surrogate-enhanced environment introduces two structural parameters: \textit{n\_envs} and \textit{buffer\_size}.

\begin{itemize}
    \item \textbf{\textit{n\_envs}}: this parameter determines the number of parallel environments managed by the RL algorithm. However, increasing the number of environments comes with a tradeoff: the policy converges more rapidly to the optimal policy, but the computation cost is vastly increased. In the PPO algorithm implemented in Stable-baselines3, rollouts are used to gather samples for policy training. These samples are stored in a structure called the \textit{ rollout\_buffer}. The policy model is updated only once the \textit{ rollout\_buffer} is full, whose capacity is defined as:  
    \[
    \textrm{rollout\_buffer\_size} = \textrm{num\_envs} \cdot \textrm{buffer\_size}.
    \]  
    As the number of parallel environments increases, the total rollout buffer size grows proportionally since the buffer size remains constant. This leads to a substantial slowdown in policy training due to the increased size of the dataset.

    \item \textbf{\textit{buffer\_size}}: in our experiments, we observed that the size of the training dataset (i.e., the number of transitions saved in the \textit{rollout\_buffer}) is not as critical for developing an effective RL policy in this environment as the frequency of policy updates. Frequent policy updates (i.e., smaller \textit{buffer\_sizes}) led to a significantly faster improvement in policy performance, even with more parallel environments. Consequently, we introduced a \textit{buffer\_size} structural parameter, which controls the number of steps per rollout before a policy update occurs. With this parameter, the policy is updated each time all parallel environments complete the defined number of steps specified by \textit{buffer\_size}. In other words, the policy is updated once the \textit{rollout\_buffer} is full.
\end{itemize}

State transitions are vectorized, enabling the surrogate model to compute transitions as follows:
\[
f: (\mathbf{S_t}, \mathbf{A_t}) \rightarrow (\mathbf{S_{t+1}}, \mathbf{r}_t),
\]
where \( \mathbf{S} \), \( \mathbf{A} \), and \( \mathbf{r}_t \) represent, respectively, the state matrix made up of \textrm{num\_envs} state vectors, the action matrix made up of \textrm{num\_envs} action vectors, and the reward vector consisting of \textrm{num\_envs} rewards.

\section{Application on Smart Grids}

\begin{figure}
    \centering
    \includegraphics[width=\linewidth]{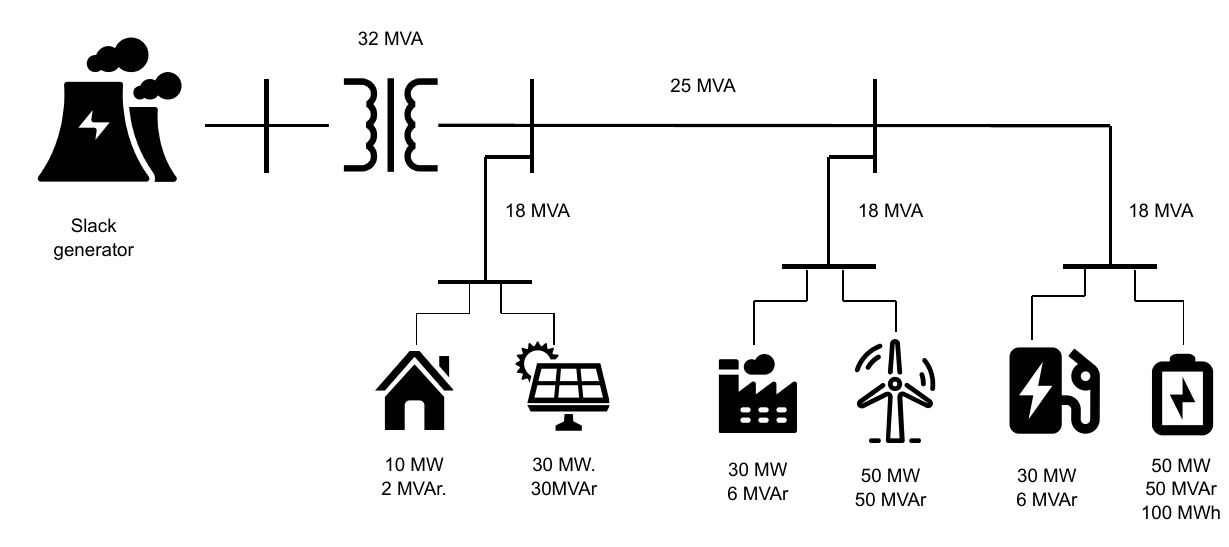}
    \caption{ANM6-Easy environment: an electrical network composed of a slack generator, $2$ renewable energy generators (a wind farm and a PV plant), $3$ passive loads (an industrial complex, a residential area, and an EV charging garage), and a storage unit.}
    \label{sgdiagram}
\end{figure}

We used the ANM6 environment in this work, which is a model of a simple distribution smart grid with a residential load with PV generation, an industrial load with wind generation, and an EV charging park with an \ac{ESS}, which are all separated in three different buses and supplied by a slack generator, a diagram of which can be seen in Figure \ref{sgdiagram}.

The main objective of the environment is to train the best policy to maintain the lowest possible energy loss in the grid due to power losses in the lines, substations, and other equipment, as well as losses due to congestion and losses due to renewable generation curtailment. It does this by controlling the power flow of the \ac{ESS} and by curtailing the renewable energy supplied by the PV and wind power plants when necessary while the rest of the energy is supplied or injected into the slack generator, representing the transmission grid.

As it can be seen, the lines that supply power to the industrial prosumer and the EV charging park have a power capacity lower than the maximum power that the loads can demand. This represents a real challenge for current distribution grids with increasing demand for loads such as electric vehicles, as they could incur expensive grid investments to accommodate higher peak power \cite{reshighcapex}, but local renewable sources or \ac{ESS}s could avoid such investments when actively controlled while providing additional benefits such as cheap energy generation and higher efficiency.

    \subsection{ANM6-Easy environment\footnote{For more details on variables definition and notation used, see \cite{anm6paper}}}


As stated in the previous section, Gym-ANM is a simulation framework used to design Reinforcement Learning environments to train policies to optimize distribution grid control systems. ANM6-Easy is a use case designed with it, representing a small distribution grid capable of simulating a wide variety of problems related to grid management \cite{anm6paper}.

The ANM6-Easy environment models a distribution network consisting of seven devices connected to six buses, where \(\mathcal{D}\) denotes the complete set of connected devices. Within \(\mathcal{D}\), there are two main subsets: \(\mathcal{D}_G\), which contains the three generators (including a slack generator), and \(\mathcal{D}_L\), which contains three loads. In addition, a distributed energy source device operates as part of the network energy storage. The slack generator is the only device connected to its bus, referred to as a slack bus, identified, for convenience, by index $1$. The slack bus serves to balance power flows and provides a voltage reference, maintaining a defined voltage of $1\,\text{p.u.}\angle 0^\circ$

Each device, bus, and branch has a set of fully known physical parameters that define its properties, behavior, and constraints \cite[see][Appendix~D]{anm6paper}.

For simplicity's sake, the environment is designed to be completely deterministic: load demands and the maximum potential generation are modeled by two fixed, 24-hour time series that repeat every day, which are assumed to be known a priori (future-awareness hypothesis).
Time is discretized with a step of $\Delta t = 15 \,\text{min}$.

At each timestep \(t\), the system’s state is represented by the vector:

\begin{equation}
\label{eq:state-vector}
\mathbf{s}_t = \left[ \{ P_{d,t} \}_{d \in \mathcal{D}}, \{ Q_{d,t} \}_{d \in \mathcal{D}}, \text{SoC}_{t} ,\{ P^{\text{(max)}}_{g,t} \}_{g \in \mathcal{D}_G - \{g^{\text{slack}}\}}, \text{aux}_t \right],
\end{equation}

where:

\begin{itemize}
    \item \(P_{d,t}\) and \(Q_{d,t}\) represent the active and reactive power injections, respectively, for each device \(d\) at time \(t\),
    \item \(\text{SoC}_{t}\) denotes the state of charge of the battery, which is the only energy storage device in the system,
    \item \(P^{\text{(max)}}_{g,t}\) specifies the maximum generation capacity for each generator $g \in \mathcal{D}_G - \{g^{\text{slack}}\}$ at time $t$,
    \item \(\text{aux}_t\) is an auxiliary time variable, included to ensure the process is Markovian. Specifically, it is an index ranging from $0$ to  $95$ ($24\,\textrm{h}/\Delta t$) used to retrieve the demand and generation capacity for the next timestep from the time series that model them.
\end{itemize}
The action vector \(\mathbf{a}_t\) at each time step \(t\) represents the control variables that the agent can adjust on various devices in the network. Specifically, \(\mathbf{a}_t\) is structured as follows:

\begin{equation}
    \label{eq:action-vector}
    \mathbf{a}_t = \left[ \{ a_{P_{g,t}} \}_{g \in \mathcal{D}_G - \{g^{\text{slack}}\}}, \{ a_{Q_{g,t}} \}_{g \in \mathcal{D}_G - \{g^{\text{slack}}\}}, a_{P_{\text{DES},t}}, a_{Q_{\text{DES},t}} \right],
\end{equation}

where:
\begin{itemize}
    \item \(a_{P_{g,t}}\) and \(a_{Q_{g,t}}\) denote the active and reactive power setpoints, respectively, for each generator \(g\) in the set of generators \(\mathcal{D}_G\), excluding the slack generator \(g^{\text{slack}}\),
    \item \(a_{P_{\text{DES},t}}\) and \(a_{Q_{\text{DES},t}}\) represent the active and reactive power adjustments for the DES, which can both inject and withdraw power to balance the network.
\end{itemize}

\subsubsection{Transition dynamics}
\label{sec:transition-dynamics}
Given the known future demand $\{P_{d, t+1}\}_{d \in \mathcal{D}_L}$, and power factor $\textit{pf}$ of each load, reactive power of loads $\{Q_{d, t+1}\}_{d \in \mathcal{D}_L}$ is calculated as:
\begin{equation}
\label{eq:loads-transition}
    Q_{d, t+1} = P_{d, t+1}\tan{(\arccos{(\textit{pf})})}\quad\forall d \in \mathcal{D}_L.
\end{equation}

For generators, constraints define feasible active and reactive power ranges, ensuring each device operates safely. The setpoints for active \(a_{P_{g,t}}\)and reactive \(a_{Q_{g,t}}\) power are mapped to the feasible set \(\mathcal{R}_{g, t}\), a convex polytope defined by generators physical parameters, network conditions, and external variables:

\begin{equation}
\label{eq:gen-feasible}
\begin{aligned}
\mathcal{R}_{g, t} = \{&(P, Q) \in \mathbb{R}^2 \mid \underline{P}_g \leq P \leq P^{\textrm{(max)}}_{g,t}, \\
&\underline{Q}_g \leq Q \leq \overline{Q}_g, \\
&Q \leq \tau^{(1)}_g P + \rho_g^{(1)}, \\
&Q \geq \tau^{(2)}_g P + \rho_g^{(2)}\},
\end{aligned}
\end{equation}

where the known coefficients \(\tau^{(1)}_g\), \(\rho^{(1)}_g\), \(\tau^{(2)}_g\), and \(\rho^{(2)}_g\) model the linear constraints imposed on the generator’s reactive power injection flexibility when operating near its maximum active power capacity.

This mapping process is a convex optimization problem, with the objective of finding the closest feasible point in \(\mathcal{R}_{g,t}\) to the chosen setpoints \((a_{P_{g,t}}, a_{Q_{g,t}})\):

\begin{equation}
\label{eq:gen-transition}
    (P_{g, t+1}, Q_{g, t+1}) = \argmin_{(P, Q) \in \mathcal{D}_{g,t} \subseteq \mathbb{R}^2} \quad \|(a_{P_{g,t}}, a_{Q_{g,t}}) - (P, Q)\|.
\end{equation}

A similar process is adopted for the DES unit, with additional constraints based on the current state of charge (SoC). The feasible set $\mathcal{R}_{\text{DES}, t}$ is:

\begin{equation}
\label{eq:des-feasible}
\begin{aligned}
\mathcal{R}_{\text{DES},t} = \{ &(P, Q) \in \mathbb{R}^2 \mid P_d \leq P \leq P_d^{\textrm{(max)}},\\ &\underline{Q}_d \leq Q \leq \overline{Q}_d, \\& Q \leq \tau_d^{(1)} P + \rho_d^{(1)}, \\& Q \geq \tau_d^{(2)} P + \rho_d^{(2)} \\& Q \leq \tau_d^{(3)} P + \rho_d^{(3)}, \\& Q \geq \tau_d^{(4)} P + \rho_d^{(4)}, \\
&P \geq \frac{1}{\eta\Delta t }(\text{SoC}_{t-1} -\overline{\text{SoC}})\\
&P \leq \frac{\eta}{\Delta t}(\text{SoC}_{t-1} -\underline{\text{SoC}})\},
\end{aligned}
\end{equation}

where \(\eta\) represents the efficiency of charging and discharging processes and

\begin{equation}
\label{eq:des-transition}
    (P_{\text{DES}, t+1}, Q_{\text{DES}, t+1}) = \argmin_{(P, Q) \in \mathcal{D}_{\text{DES},t+1} \subseteq \mathbb{R}^2} \quad \|(a_{P_{\text{DES},t}}, a_{Q_{\text{DES},t}}) - (P, Q)\|.
\end{equation}

The state of charge for each DES is then updated as:

\begin{equation}
\label{eq:soc-transition}
    \text{SoC}_{t+1} = \begin{cases}
        \text{SoC}_{t} - \eta \Delta t P_{\text{DES}, t+1} &\ \text{if $P_{\text{DES}, t+1} \leq 0$} \\
        \text{SoC}_{t} - \frac{\Delta t}{\eta} P_{\text{DES}, t+1} &\ \text{otherwise}.
    \end{cases}
\end{equation}

With active and reactive power known for each device (except for the slack generator), simply grouping and summing for each bus, we get $\{P_{i, t+1}^{(\text{bus})}\}_{i=2}^6$ and $\{Q_{i, t+1}^{(\text{bus})}\}_{i=2}^6$. Along with slack bus voltage definition and admittance of branches $Y_{ik} = G_{ik}+jB_{ik}$, it is possible to solve power flow equations:\footnote{$\theta_{ik} = \theta_i - \theta_k$}
\begin{align}
\label{eq:active-power-balance}
 P^{\text{(bus)}}_{i, t+1} &= \sum_{k=1}^6|V_{i, t+1}||V_{k, t+1}|(G_{ik}\cos\theta_{ik, t+1} + B_{ik}\sin\theta_{ik, t+1})\\Q^{\text{(bus)}}_{i, t+1} &= \sum_{k=1}^6|V_{i, t+1}||V_{k, t+1}|(G_{ik}\sin\theta_{ik, t+1} + B_{ik}\cos\theta_{ik, t+1}),\label{eq:reactive-power-balance}
\end{align}
obtaining $V_{i, t+1} = |V_{i, t+1}|\angle\theta_{i, t+1}$ and active and reactive power of the slack generator.

Finally, it is straightforward to determine the directed branch current and apparent power flows:\footnote{For details, see \cite{anm6paper}}

\begin{equation}
    \label{eq:branch-current}
    \begin{pmatrix}
        I_{ij, t+1}\\
        I_{ji, t+1}
    \end{pmatrix} =
    \begin{pmatrix}
        \frac{1}{|t_{ij}|^2}(y_{ij} + y_{ij}^{\text(sh)})&-\frac{1}{|t^*_{ij}|^2}y_{ij}\\
        -\frac{1}{|t_{ij}|^2}y_{ij}&(y_{ij} + y_{ij}^{\text(sh)})
    \end{pmatrix}
    \begin{pmatrix}
        V_{i, t+1}\\
        V_{j, t+1}
    \end{pmatrix}
    \end{equation}
    \begin{align}
        \label{eq:apparent-power-flow-1}
        |S_{ij, t+1}|&= |V_{i, t+1}I_{ij, t+1}^*|\\
        |S_{ji, t+1}|&= |V_{j, t+1}I_{ji, t+1}^*|.\label{eq:apparent-power-flow-2}
    \end{align}

\subsubsection{Reward function}
The reward takes into account energy losses and a penalty term for violation of operating conditions:

\begin{equation}
    \label{eq:reward}
    r_t = -\left(\Delta E_{t:t+1} + \lambda \Phi(\mathbf{s}_{t+1})\right).
\end{equation}

The energy loss term is composed of three sub-terms that take into account the main types of energy loss sources that this type of smart grid can have, which are: 
\begin{itemize}
    \item Transmission losses due to energy leaks in substations and lines.
    \item The net energy that flows from the grid to the DES unit, which approximates to the losses due to battery inefficiencies \cite{anm6paper}.
    \item Losses due to curtailment of renewable energy, which occurs when the grid cannot accept all of the energy that the generator can provide due to lack of demand, line congestion, or other discretional decisions taken by the grid operator.
\end{itemize}
In mathematical terms:
\begin{align}
    \label{eq:energy-loss-1}
    \Delta E_{t:t+1} &= \Delta E^{(1)}_{t:t+1} + \Delta E^{(2)}_{t:t+1}+ \Delta E^{(3)}_{t:t+1}\\\label{eq:energy-loss-2}
    \Delta E^{(1)}_{t:t+1} &= \Delta t \sum_{d \in \mathcal{D}}P_{d, t+1}\\\label{eq:energy-loss-3}
    \Delta E^{(2)}_{t:t+1} &= -\Delta t P_{\text{DES}, t+1}\\
    \Delta E^{(3)}_{t:t+1} &= \Delta t \sum_{g \in \mathcal{D}_g - \{g^{\text{(slack)}}\}}\left(P^{\text{(max)}}_{g, t+1} - P_{g, t+1}\right).
\end{align}

The penalty function considers two specific grid constraints: the rated power of the lines and substations and the voltage limit of the buses. These constraints prevent overheating of the grid equipment and ensure voltage stability to the devices connected to the grid.
\begin{multline}
\label{eq:penalty}
    \Phi(\mathbf{s}_{t+1}) = \Delta t\bigg(\sum_{i = 1}^6 \left(|V_{i, t+1}| - \overline{V}_i\right)^+ +\left(\underline{V}_i - |V_{i, t+1}|\right)^+ \\+\sum_{i, j}\left(|S_{ij, t+1}| - \overline{S}_{ij}\right)^+ +\left(\underline{S}_{ij} - |S_{ij, t+1}|\right)^+\bigg).
\end{multline}

In our environment $\lambda = 100$ and, for numerical stability, $r_t$ is clipped to the range $[-100, 100]$.

\subsubsection{Terminal state}\label{sec:terminal_state}
A terminal state is defined as a state with no solution for \eqref{eq:active-power-balance} and \eqref{eq:reactive-power-balance} given the chosen actions, indicating a collapse of the grid, which typically occurs when the voltage constraints cannot be met. This results in a voltage dip, which is a rather common type of fault regarding power quality management and has undesirable consequences mainly in industries \cite{pqprobs}.

    \subsection{Surrogate model}
    \label{sec:surrogate}
    The proposed surrogate for the ANM6-Easy environment is designed to simulate state transitions within a Markov Decision Process (MDP), predicting the future state of the system and the reward associated with the transition:

    \begin{equation}\label{eq:surrogate}
        (\mathbf{s}_{t+1}, r_t) = \text{Surrogate}(\mathbf{s}_{t}, \mathbf{a}_{t}).
    \end{equation}
    This surrogate model is specifically built to achieve these predictions, speeding up the computation and making it much faster than the transitions occurring within the original environment.

The surrogate consists of three neural networks, each of which models different parts of the environment’s dynamics: 
\begin{itemize}
\item power transitions of no-slack generators, described by \eqref{eq:gen-transition}
\item power transitions of the battery, described by \eqref{eq:des-transition}, 
\item bus voltages and power transition of the slack generator, described by \eqref{eq:active-power-balance} and \eqref{eq:reactive-power-balance}.
\end{itemize}
The three parts of the model are described below.

\subsubsection{Generators state}
\label{sec:gen-net}
As explained in Subsection \ref{sec:transition-dynamics}, the mapping from $(a_{P_{g,t}}, a_{Q_{g,t}})$ to $(P_{g, t+1}, Q_{g, t+1})$ consists of solving the convex optimization problem \eqref{eq:gen-transition}.

That problem could be stated in standard form as:
\begin{equation}
\label{eq:gen-opt-prob}
\begin{aligned}
\min_{(P_{g, t+1}, Q_{g. t+1})} \quad &\|(a_{P_{g,t}}, a_{Q_{g,t}}) - (P_{g, t+1}, Q_{g. t+1})\| \\
\textrm{s.t.}\quad & G (P_{g, t+1}, Q_{g. t+1}) - h \leq 0\\
\end{aligned}
\end{equation}

\begin{equation}
G = \begin{pmatrix}
-1 & 1 & 1 & 0 & 0 & -\tau^{(1)}_g & \tau^{(2)}_g\\
0 & 0 & 0 & -1 & 1 & 1 & 1
\end{pmatrix}^T
\end{equation}

\begin{equation}
h = \begin{pmatrix}
\underline{P}_g & \overline{P}_g & P^{\textrm{(max)}}_{g,t} & \underline{Q}_g & \overline{Q}_g & \rho^{(1)}_g & -\rho^{(2)}_g
\end{pmatrix}^T.
\end{equation}
Being \eqref{eq:gen-opt-prob} a convex problem whose constraints satisfy Slater's condition, $(P_{g, t+1}, Q_{g. t+1})^*$ is an optimal solution for \eqref{eq:gen-opt-prob} if and only if there exist $\lambda^* \in \mathbb{R}^7$ such that the Karush-Kuhn-Tucker (KKT) conditions are satisfied.

Following \cite{femine2024kktinformed}, a KKT-Informed Neural Network is defined to predict $\{(P_{g, t+1}, Q_{g. t+1})^*\}_{g \in \mathcal{D}_g - \{g^{\text{slack}}\}}$ having as input $\{ a_{P_{g,t}}, a_{Q_{g,t}}, P^{\textrm{(max)}}_{g,t} \}_{g \in \mathcal{D}_G - \{g^{\text{slack}}\}}$.

\subsubsection{Battery state}
\label{sec:des-net}
Analogously, \eqref{eq:des-transition} could be stated as:
\begin{equation}
\label{eq:des-opt-prob}
\begin{aligned}
\min_{(P_{\text{DES}, t+1}, Q_{\text{DES}. t+1})} \quad &\|(a_{P_{\text{DES},t}}, a_{Q_{\text{DES},t}}) - (P_{\text{DES}, t+1}, Q_{\text{DES}. t+1})\| \\
\textrm{s.t.}\quad & G (P_{\text{DES}, t+1}, Q_{\text{DES}, t+1}) - h \leq 0\\
\end{aligned}
\end{equation}

\small
\begin{equation}
G = \begin{pmatrix}
-1 & 1  & 0 & 0 & -\tau^{(1)}_\text{DES} & \tau^{(2)}_\text{DES}&\tau^{(3)}_\text{DES}&\tau^{(4)}_\text{DES}&-1&1\\
0 & 0 & -1 & 1 & 1 & 1 & 1 & 1 & 0 & 0
\end{pmatrix}^T
\end{equation}
\normalsize
\begin{equation}
\begin{split}
h &= \left(\begin{matrix}
\underline{P}_\text{DES} & \overline{P}_\text{DES} &  \underline{Q}_\text{DES} & \overline{Q}_\text{DES} \end{matrix}\right.\\&\qquad\
\begin{matrix}\rho^{(1)}_\text{DES}&-\rho^{(2)}_\text{DES}&-\rho^{(3)}_\text{DES}&\rho^{(4)}_\text{DES}\end{matrix}\\&\qquad\left.\begin{matrix}\frac{1}{\eta\Delta t }\left(\text{SoC}_{t} -\overline{\text{SoC}}\right), \frac{\eta}{\Delta t}\left(\text{SoC}_{t} -\underline{\text{SoC}}\right)\end{matrix}\right)^T.
\end{split}
\end{equation}
Therefore, a KKT-Informed Neural Network could be trained to predict $\left(P_{\text{DES}, t+1}, Q_{\text{DES}. t+1}\right)^*$ having as input $\left[ a_{P_{\text{DES},t}}, a_{Q_{\text{DES},t}},\text{SoC}_{t}\right]$.

\subsubsection{Power balance}
\label{sec:power-balance-net}
With active and reactive power determined for each device (except for the slack generator), these values are grouped by their associated buses, yielding $P^{\text{(bus)}}_i$, $Q^{\text{(bus)}}_i$  $i = 2, \dots, 6$ (except for the slack bus, identified by $i=1$).

The power balance network uses these aggregated powers as input to predict bus voltages $\{|V_i|, \theta_i\}_{i=2}^6$. Since the slack bus voltage is fixed ($V_1 = 1\,\text{p.u.}$ and $\theta_1 = 0^{\circ}$),  it is possible to calculate both sides of \eqref{eq:active-power-balance},  \eqref{eq:reactive-power-balance} for $i = 2, \dots, 6$. The network is trained to minimize discrepancies between these two terms, effectively learning to solve power balance equations. In particular, the following per-sample loss will be minimized during the training:

\begin{align}
    \label{eq:loss-power-balance}
    &\mathcal{L} = \mathcal{L}_1 + \mathcal{L}_2\\
    &\mathcal{L}_1 = \frac{1}{5}\sum_{i=2}^6\left( P^{\text{(bus)}}_i - \sum_{k=1}^6|V_i||V_k|\left(G_{ik}\cos\theta_{ik} + B_{ik}\sin\theta_{ik}\right)\right)^2\\
    &\mathcal{L}_2 = \frac{1}{5}\sum_{i=2}^6\left(Q^{\text{(bus)}}_i - \sum_{k=1}^6|V_i||V_k|(G_{ik}\sin\theta_{ik} + B_{ik}\cos\theta_{ik})\right)².
\end{align}

Once $\{|V_i|, \theta_i\}_{i=1}^6$ are known, the active and reactive powers for the slack generator can be calculated:

\begin{align}
    \label{eq:active-power-slack-generator}
    P^{\text{(bus)}}_1 &= \sum_{k=1}^6|V_1||V_k|(G_{1k}\cos\theta_{1k} + B_{1k}\sin\theta_{1k})\\Q^{\text{(bus)}}_1 &= \sum_{k=1}^6|V_1||V_k|(G_{1k}\sin\theta_{1k} + B_{1k}\cos\theta_{1k}).\label{eq:reactive-power-slack-generator}
\end{align}

\subsubsection{Next state}

Under the future-awareness assumption, $\{P_{d, t+1}\}_{d \in \mathcal{D}_L}$ and $\{ P^{\text{(max)}}_{g,t+1} \}_{g \in \mathcal{D}_G - \{g^{\text{slack}}\}}$ are already known. Via the networks defined in Sections \ref{sec:gen-net}, \ref{sec:des-net}, \ref{sec:power-balance-net}, equations \eqref{eq:loads-transition} and \eqref{eq:soc-transition} and the relation
\begin{equation}
    \text{aux}_{t+1} = (\text{aux}_{t} + 1) \bmod{96}
\end{equation}
that ensures daily periodicity,
all information to build $\mathbf{s}_{t+1}$, by definition \eqref{eq:state-vector}, is available.
\subsubsection{Reward}

All information to compute \eqref{eq:branch-current}-\eqref{eq:penalty} is available.

\section{Results and discussion}\label{results}
\subsection{Surrogate training conditions}
Each of the three constituent modules of the surrogate is defined as a multilayer perceptron with three hidden layers of $512$ neurons each, a residual connection between the inputs and outputs of the latter, and with a LeakyReLU (slope of $0.01$) as the activation function. 

Each module is optimized with AdamW, with a learning rate of 1e-5. An early stopping condition is posed: training stops with no progress in the last 5000 steps. 

At each training step, a batch of $64$ samples is sampled from a Sobol sequence of dimensionality $21$.

All dimensions are scaled to provide inputs for each part of the surrogate, as outlined in \ref{sec:surrogate}. For more information on the scaling process, refer to \ref{sec:scaling}.

Our proposed \ac{PINN}-based surrogate model has been analyzed 
 against other methods from the state-of-the-art, considering seven types of surrogates trained from two different kinds of datasets. The studied surrogates are Deep Neural Networks (DNN), XGBoost (XGB), Decision Trees (DT), Random Forest (RF), Linear Regressor (LR), and our proposed method with \ac{PINN}s. 

These models (except for the \ac{PINN}) were trained using two types of datasets: Generative and Agent-based. Following the naming convention established in \cite{cestero2024building}, a Generative dataset is one constructed by sampling transitions across the state space using a general sampling method---in this case, using Sobol sampling \cite{burhenne2011sampling}. In contrast, an Agent-based dataset is derived from the natural interaction of a specifically designed agent with the environment, represented here by a Random agent. Both datasets consist of $100\,000$ samples.

The loss function used in the training process is the \ac{MAE} for each model. However, the coefficient of determination metric (R$^2$) has been used to evaluate the performance of the trained models.

The results reported in the following sections have been obtained by training the models on an Intel Core i5-9499F CPU.

\subsection{Surrogate training results}

This section presents the results regarding the surrogates' training metrics. 

\begin{figure}[tbp]
                
                \centering
                \includegraphics[width=\linewidth]{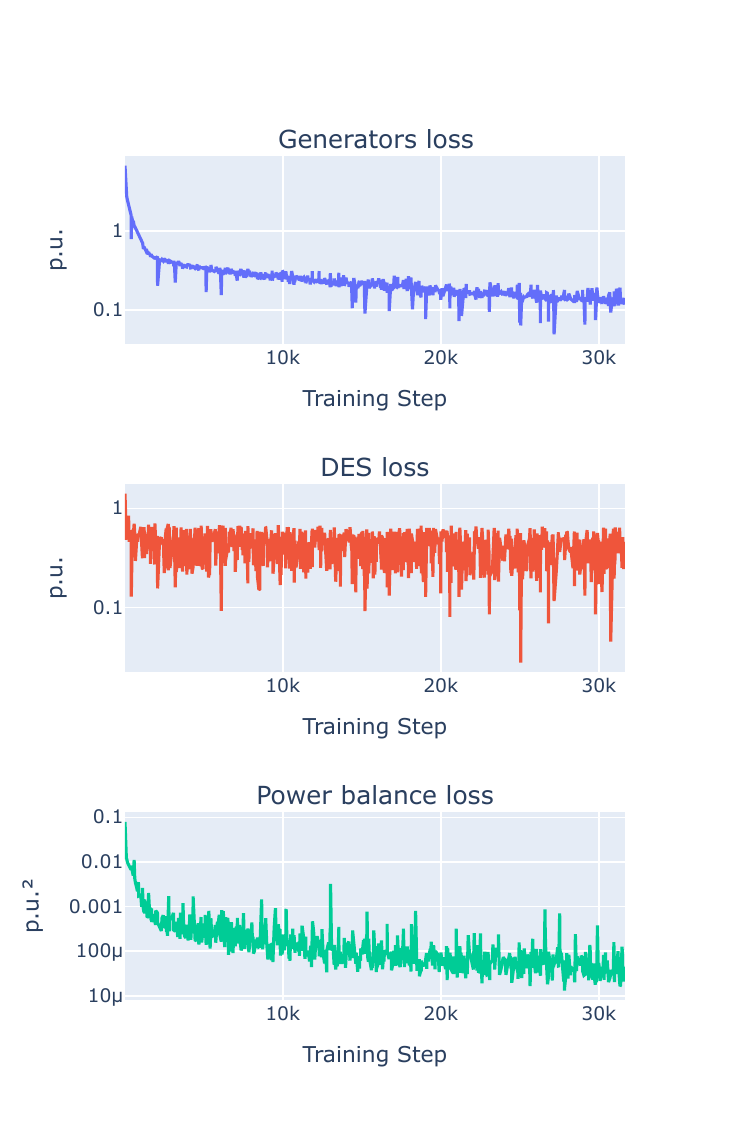}
                \caption{\textbf{Generators Loss}: Aggregate of KKT residuals \cite{femine2024kktinformed} associated with generator-related KKT-NN. \textbf{DES Loss}: Aggregate of KKT residuals \cite{femine2024kktinformed} associated with battery-related KKT-NN. \textbf{Power Balance Loss}: Residuals from power balance equations \eqref{eq:loss-power-balance} associated with voltage-related \ac{PINN} }
                \label{fig:training-loss}
            \end{figure}

Figure \ref{fig:training-loss} shows trends of losses for generators and DES network, as defined in \cite{femine2024kktinformed}, and loss \eqref{eq:loss-power-balance} for power-balance solver network, over the training.

\begin{figure}[tbp]    
                \centering
                \includegraphics[width=\linewidth]{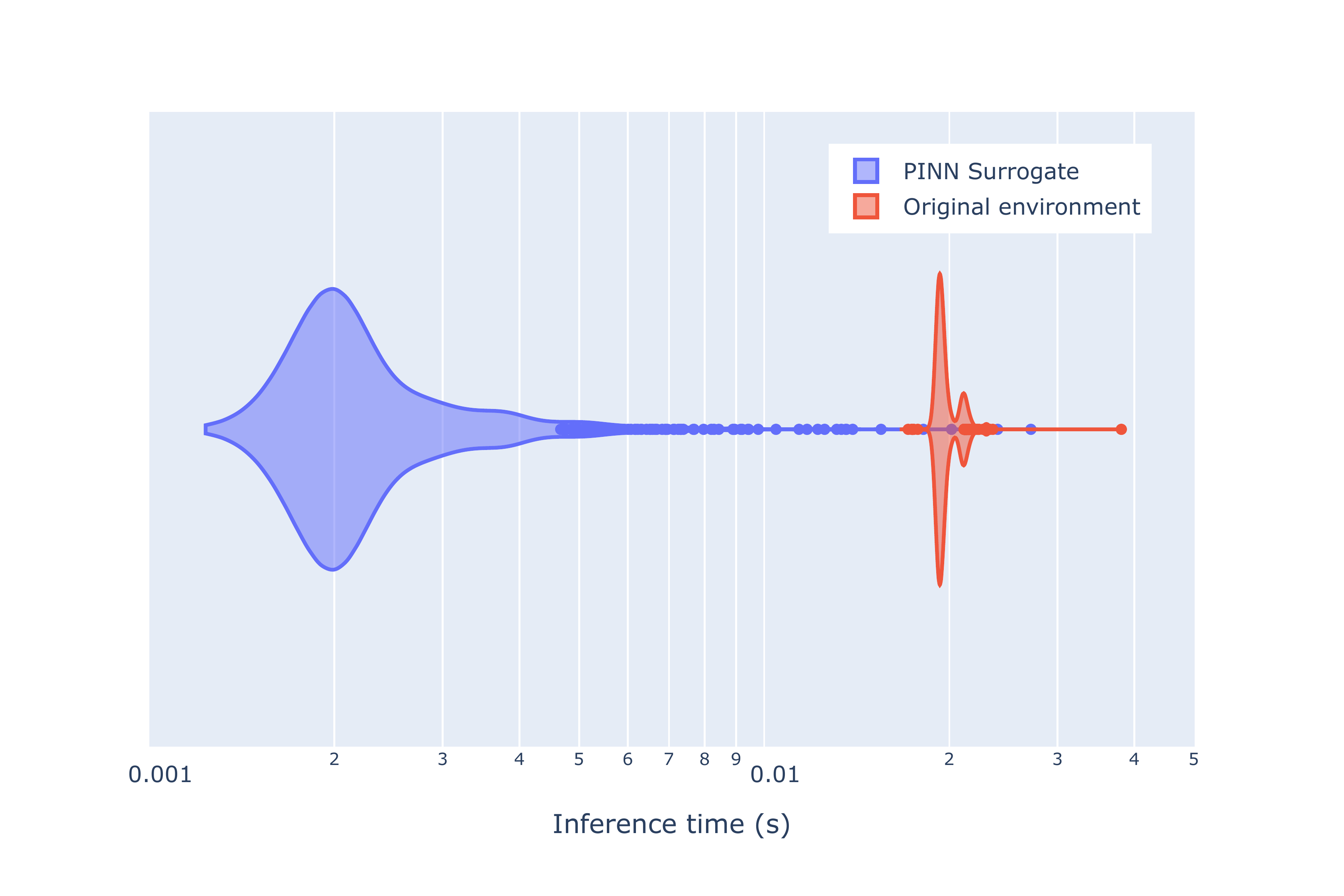}
                \caption{Comparison of inference times over 1000 transitions. The median time for \ac{PINN} stands at $0.0021~\text{s}$, while the median for the original environment is $0.0194~\text{s}$.}
                \label{fig:inference-time}
            \end{figure}

Additionally, the speed increase of the surrogate compared to the original environment is calculated by performing 1000 transitions with both. As shown in Figure \ref{fig:inference-time}, there is a median boost of almost 10 times.
            

\begin{figure}[tbp]
    \centering
    \includegraphics[width=1\linewidth]{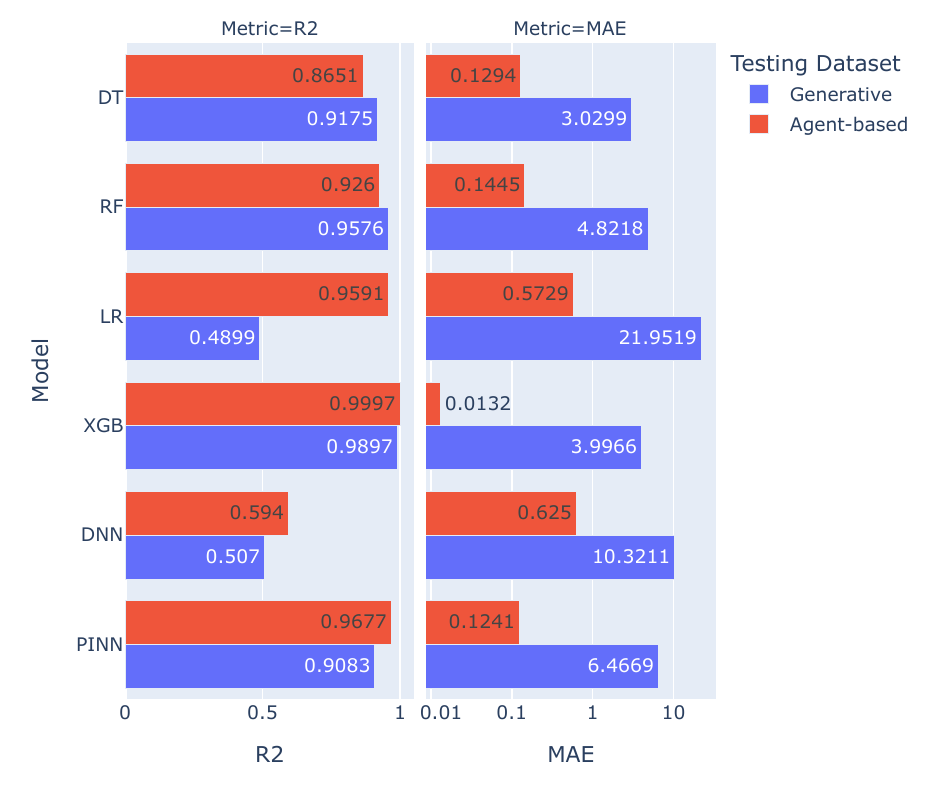}
    \caption{R$^2$ and MAE score comparison between the studied surrogate models over the two different datasets: Generative and Agent-based transitions. The R$^2$ is in linear scale while the \ac{MAE} is in logarithmic scale. Clearly, the XGB models show the highest accuracy and the smallest error, followed closely by the PINN model. However, despite their good overall accuracy, most surrogates do not perform well when used as an RL training environment, excluding the PINN model.}
    \label{fig:surrogate_accuracy_comparison}
\end{figure}

Figure \ref{fig:surrogate_accuracy_comparison} shows the accuracy comparison between all the studied models in terms of R$^2$ and \ac{MAE}. In the same line, as the conclusions of \cite{cestero2024building}, in general, these results show that Agent-based sampling produces better performance than Generative sampling in most cases. This is clearly seen in the \ac{MAE} metric, which is in logarithmic scale, and the error of the models trained in the generative dataset is generally much higher than the other one.

These results also show that the model with the highest accuracy is XGB, followed closely by the \ac{PINN} model. From these results, we could expect the XGB and \ac{PINN} models to perform best as a proper environment for \ac{RL} training, while the models DNN and LR may have the least favorable performance. However, this is not what happens in practice, as further results reveal. Once the models operate outside their training datasets, their performance shifts significantly.

\begin{figure}[hbp]
    \centering
    \includegraphics[width=1\linewidth]{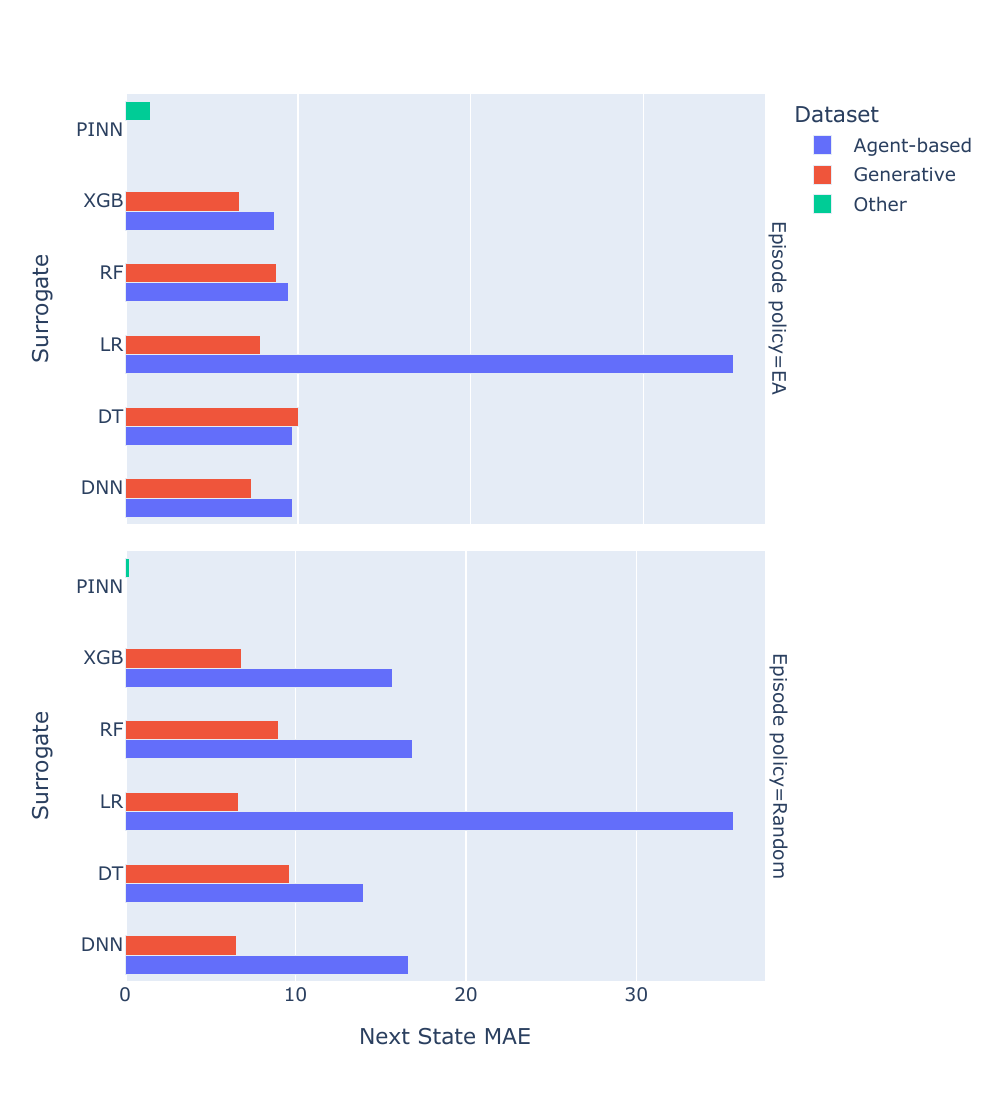}
    \caption{MAE of the different models, divided by their training datasets (Agent-based or Generative) averaged during a random episode using two different policies: Expert Agent (EA) and a Random policy. The results indicate that the \ac{PINN} model shows the least error. Dataset \textit{Other} means that no dataset has been used for training the surrogate.}
    \label{fig:episodic_MAE_comparison}
\end{figure}

The overall accuracy of the models in terms of R$^2$ suggests that many models, in principle, should be able to represent the transitions of the environment accurately. To validate this, we run a random episode using two different agents: a pre-trained agent that is able to follow the optimal policy in the environment, called \ac{EA}, and a random agent whose actions are chosen with a uniform probability from the action space of the environment. We then calculate the \ac{MAE} of all the surrogates throughout the episode of each agent. Figure \ref{fig:episodic_MAE_comparison} shows the averaged overall MAE of each surrogate model. As per the figure, our method shows the least error, even though Figure \ref{fig:surrogate_accuracy_comparison} shows that other models display more accurate results. Nevertheless, when actually representing a realistic episode, the surrogate model trained using \ac{PINN}s presents itself as the most suitable method. This means that purely data-driven surrogate models fail to extrapolate outside their training datasets, while the \ac{PINN} model is reliable across the entire state space.

\begin{figure*}[t]
    \centering
    \includegraphics[width=1\linewidth]{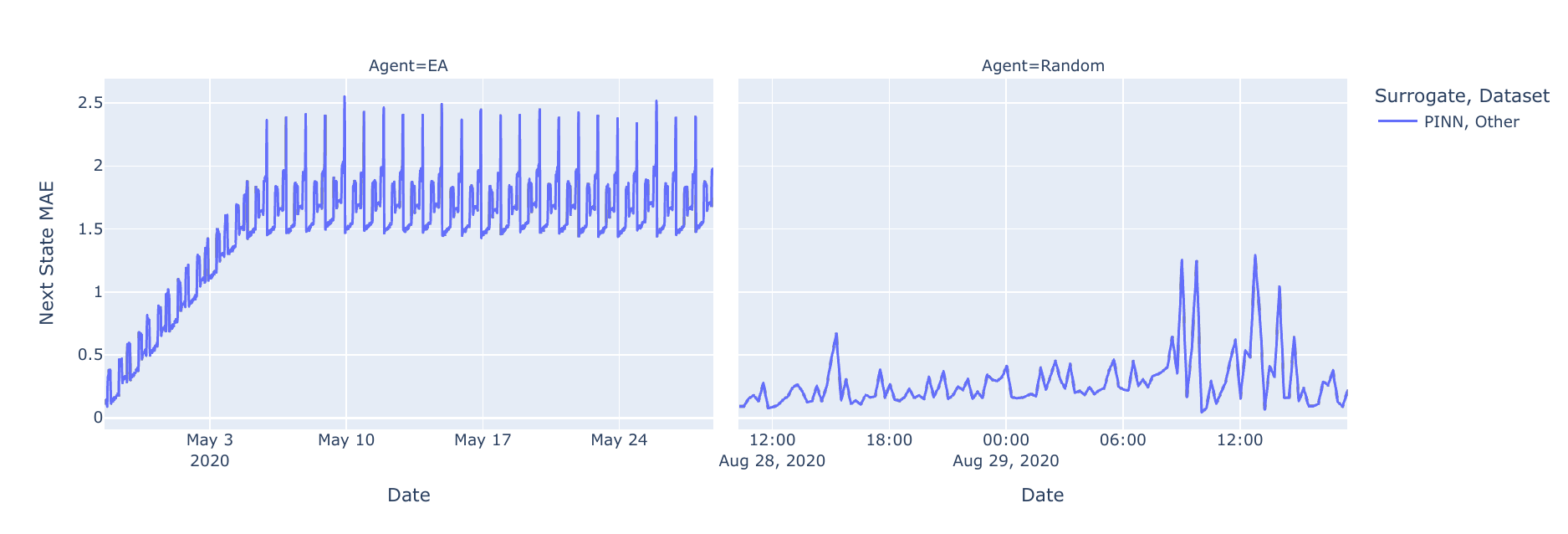}
    \caption{MAE of the \ac{PINN} agent during a random episode driven by an Expert Agent (EA) and a Random Agent. The EA results show a daily error periodicity and a linear increase of the average error until stability is reached. This is due to the nature of the environment: the stability corresponds to reaching optimal stability values of the load of the battery. On the other hand, the errors shown by the Random agent do not show clear periodicity even though the overall magnitude of the error is smaller.}
    \label{fig:episode_PINN_MAE}
\end{figure*}

Figure \ref{fig:episode_PINN_MAE} depicts the evolution of the MAE of the \ac{PINN} surrogate during episodes driven by an expert and a random agent. The random episode does not show any interesting behavior despite the low error during almost the entire episode. Noticeable features of the results include the daily periodicity of the episode and the two different stages of the episode --- its initial \textit{slope} and subsequent stability. 


We reproduced a render of this simulation and saw that the periodicity shown in these figures happens due to two factors. One is the battery, which slowly charges from zero with charging and discharging oscillations, albeit with a linear daily mean growth, until it reaches a steady state. The second one is the increase in energy demand from the loads. In particular, electric vehicles have a demand peak during the last hours of the day, and the lower end of the error period corresponds to the hours of the night, where the demand is lowest. In any case, the error of the \ac{PINN} surrogate is still noticeably lower than for the rest of the algorithms.


\subsection{Optimization of environment structural parameters}

We exploit the parallelizable nature of these surrogate models as stated previously, by predicting several transitions at the same time, thus having several environments at the same time. Nevertheless, using too many parallel environments may lead to saturation in the training speed and a decay of the results.

We found that the PPO algorithm from Stable-Baselines 3 becomes impaired during the network training step, as the network is overwhelmed with an excessive number of samples, hindering efficient training. Therefore, we found that by changing the rollout buffer size of the algorithm, the training speed was greatly affected. However, decreasing the rollout buffer has some drawbacks because if the buffer size is smaller than one episode, it may lead the policy to unstable results.

\begin{figure}[b!h]
    \centering
    \includegraphics[width=1\linewidth]{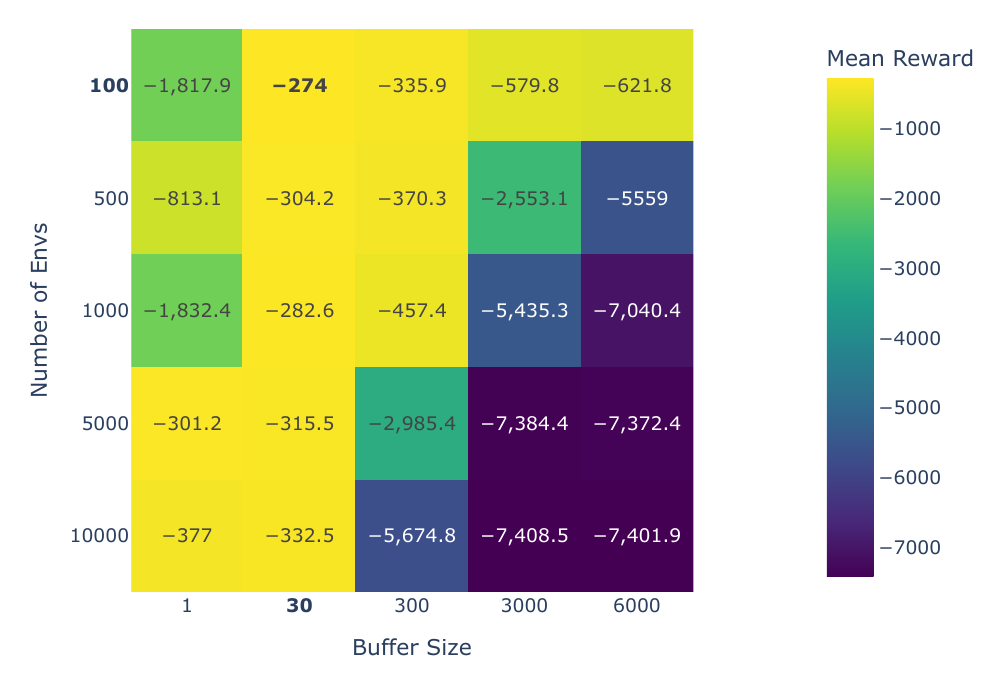}
    \caption{Results of the structural parameter optimization process in terms of reward acquired by each combination, averaged to the last 10 evaluations. Some results were cut to a maximum number of steps dependent on the number of training steps performed or if they arrived at convergence. The real buffer size of the rollout buffer is a product of the number of environments and the hyperparameter shown in this figure}
    \label{fig:HP_optimization_1}
\end{figure}

We analyze many combinations of these two variables and arrive at the best combination of these parameters, shown in Figure \ref{fig:HP_optimization_1}. Taking into account that one episode in this environment contains $3\,000$ steps, all the columns with smaller buffer sizes do not reach finishing one episode. However, the figure shows that the policy does not need as much. The best combination of structural parameters shown in this analysis is \textbf{100 environments} and a \textbf{buffer size of 30}.

This analysis uses some early stopping approaches to finish the experiment after arriving at plateau values (not increasing the best value after several evaluations) or after reaching an arbitrary number of training steps, dependent on the number of evaluations of the corresponding experiment.

\begin{figure}[tbp]
    \centering
    \includegraphics[width=1\linewidth]{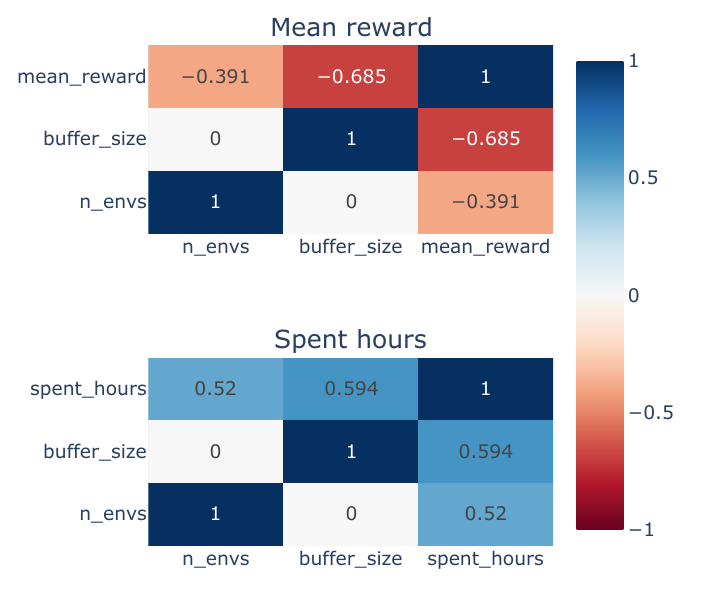}
    \caption{Correlation of the structural parameters buffer size and number of environments with the mean reward and the spent time. The results show an inverse correlation of the reward with the buffer size and the number of environments, with the former being the highest one; and in terms of spent time, the parameters show a similar correlation.}
    \label{fig:HP_correlation_reward}
\end{figure}

Analyzing the results of structural parameter optimization more deeply, we calculated the correlation of these parameters with the mean reward of the episode and the time spent training. We present the results of this analysis in Figure \ref{fig:HP_correlation_reward}. As depicted, a representative inverse correlation between the mean reward of the episode and the buffer size can be seen, showing the effect that we saw in the early phases of the parallelization trials with the environment. In addition, it can be seen that the effect of the number of environments is not as impactful as the buffer size. In contrast to the time correlations, it can also be seen that the time to finish training is correlated proportionately to the buffer size and the number of environments, but both variables, in this case, have almost the same effect on the total training time. 

\subsection{Discussion}
\begin{figure}[tbp]
    \centering
    \includegraphics[width=1\linewidth]{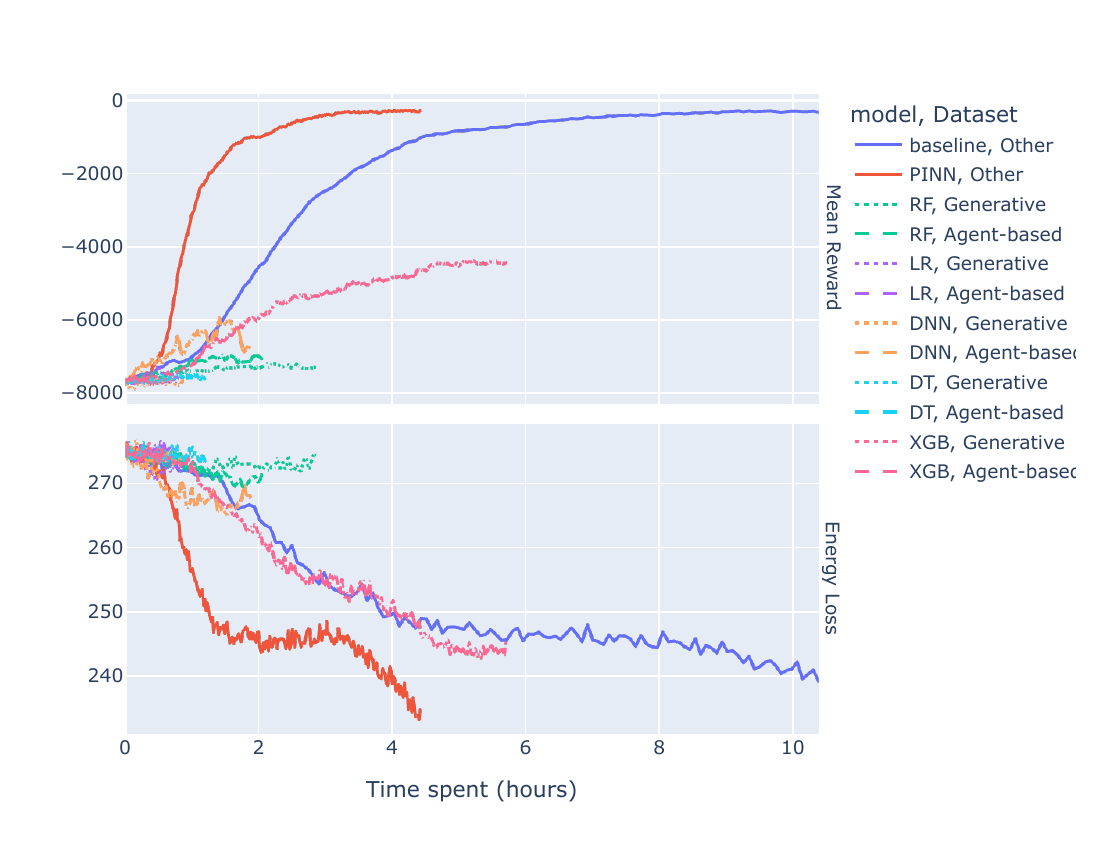}
    \caption{Reward evolution over spent time for all the surrogate models and the baseline (real simulator). The results show a great increase in performance using the \ac{PINN} surrogate over the raw simulator. Also, it is shown that the other surrogates are incapable of converging to a good policy due to their poor modeling of the inherent physical nature of the simulator.
    The evolution of energy loss is shown below. The \ac{PINN} surrogate is able to converge faster to a solution that saves the highest amount of energy.}
    \label{fig:final_results}
\end{figure}

Using these optimized structural parameters, we used all the studied models as a surrogate environment to train a PPO agent to reach an optimal policy. Figure \ref{fig:final_results} shows the evolution of the agent's training reward for each surrogate model, as well as the baseline, which shows the evolution of the training of a PPO agent using the original environment without any surrogate models. All the agents were limited by an early stopping condition that stopped the training process after reaching saturation (not being able to increase the reward after $20$ episodes)

The results show that our method using a surrogate model trained using \ac{PINN}s is much faster than the baseline, achieving a similar score. Moreover, it is shown that the \ac{PINN} surrogate is the only one capable of achieving a working policy that is far from being random. This is achieved because the \ac{PINN} surrogate is the only one capable of properly modeling the physics behind the original environment, and it is not limited to replicating the environment's behavior blindly. 

In terms of energy loss, the Smart Grid operated by the \ac{PINN} agent is able to decrease the energy loss of the network faster and more reliably than all of the other models, even the baseline environment. Observing some simulation renders, it has been found that the agent is capable of properly modeling the main causes of inefficiencies and power losses, especially regarding the battery, which operates consistently between 80\% and 20\% of charge, which are its most efficient thresholds in terms of power losses and also the health of the battery.

Furthermore, the line that supplies the bus connected to the battery and EV chargers has a rated power lower than the maximum power demanded by the EVs, and despite this, the agent decides to exploit the battery mainly to cover the power required by the EV that cannot be physically supplied by the line, which shows the degree of detail in the modeling of the physical needs of the loads and resources connected to the grid. 

Another observation made is that the agent decides to curtail most of the power that the renewable generators can provide, even if it implies an added power loss and results in power demanded from the slack generator. Possibly, this is the best balance that the agent can identify between other power losses and voltage penalties, although it might result in lower cost performance due to the energy generated from the slack generator.

Finally, looking at the results in Figure \ref{fig:surrogate_accuracy_comparison}, one may expect the XGB model to be the best one since it has almost a perfect R$^2$. However, even though the accuracy of the transitions seems to be high, in long-term simulations, the XGB model accumulates a prediction error, which causes an RL agent to fail to acquire in a robust way the knowledge to replicate a realistic policy, which is not a problem with the \ac{PINN} surrogate. Our model is able to understand deeply how the environment works, following the physics of the state transitions, which means a reduced error accumulation over the state transitions and predicting realistic transitions even if in unexplored regions of the state space. This can be clearly seen in Figure \ref{fig:episodic_MAE_comparison}.

From a practical application point of view, this means that with a \ac{PINN} surrogate, thanks to its capability to extrapolate their prior physical knowledge, it is possible to drastically change the demand and generation profiles of the devices connected to the grid while maintaining the same guarantees on accuracy and safety, without having to re-train the model. In real distribution networks, loads frequently change their typical demand patterns for a number of reasons, such as consumers buying new equipment or a change of tenants that have different consumption needs, and distributed generators can increase their rated power while using the same point of connection. Being able to perform reliable simulations with these new patterns without having to re-train has been identified as a clear advantage regarding the time and computing demand that \ac{PINN}s offer with respect to other agents.

\section{Conclusions}\label{Conclusions}

This work has demonstrated the efficacy of using \ac{PINN}s to build surrogate models for optimizing energy management in Smart Grids through RL. Using the Gym-ANM framework, we successfully simulated a simple Smart Grid environment and trained an RL agent to optimize energy management in the grid.

Our novel approach, which combines \ac{PINN}s with surrogate modeling, has shown significant improvements in training efficiency compared to traditional methods. Specifically, we achieved a 50\% reduction in the policy training time and a 10 times speed up in the inference time while maintaining the accuracy and reliability of the RL policies for grid management, while classic surrogate models were unable to converge to a reliable policy. This enhancement in computational efficiency is crucial for addressing the increasing complexity of modern Smart Grids, particularly in the context of integrating renewable energy sources and managing dynamic consumption patterns.


The results of this work have important implications for the future of Smart Grid management and optimization. Principally, the reduction of the training/inference time with negligible errors provides a clear advantage in terms of time and computing demands.

The ability of a \ac{PINN} surrogate model to accurately simulate the grid, even with a drastic change in the demand and generation of the devices connected to the grid, is a great advantage in terms of applicability to smart grid operation with respect to other RL agents. Furthermore, a clear modeling of the physical conditions of the grid allows for a safe and efficient operation, which results in lower operating and maintenance costs due to equipment fatigue and substitution.

As such, by demonstrating the potential of \ac{PINN}-based surrogate models in \ac{RL} applications, we have potentially eased the path for more rapid, efficient, cost-effective, and sustainable development of smart grid management strategies.

Future work should focus on scaling this approach to larger and more complex grid systems, as well as investigating its applicability to other aspects of Smart Grid management, such as adding demand response strategies, considering energy pricing, and integrating a wider range of renewable energy sources. In addition, other scenarios with realistic weather conditions and energy pricing objectives may provide substantial insights into the limits of automatic optimization via RL of this environment.

Our work contributes to the ongoing efforts to enhance the efficiency, reliability, and sustainability of Smart Grids. By bridging the gap between physics-based modeling and machine learning techniques, we have presented a promising approach for tackling the challenges of modern energy systems in an era of increasing complexity and environmental concerns.

\section*{Funding}

This research has been supported by the Spanish Ministry (NextGenerationEU Funds) through Project IA4TES (Grant Number: MIA.2021.M04.0008).

This project was achieved within the framework of the BEACON project (File: KK-2023/00085), submitted under the 2023 call for Collaborative Research Grants in Strategic Areas – Elkartek Program.

\appendix

\section{Terminal classification model for the surrogate environment}
\label{sec:app_terminal_classificator}

The dataset used to train this model was obtained using different realistic trajectories from the original environment. After accumulating several episodes, we balanced the data, taking all the one-state transitions that arrived at terminal states along with randomly selected non-terminal data, reducing the dataset to $137\,131$ values to classify into terminal and non-terminal states.

After comparing different types of classifiers, we opted to use a classifier based on XGBoost \cite{Chen_2016} to obtain the most accurate model possible. We also used Bayesian optimization \cite{snoek2012practical} to search for the most suitable hyperparameters:
\begin{itemize}
    \item \textit{n\_estimators}: $2$,
    \item \textit{colsample\_bytree}: $1$,
    \item \textit{learning\_rate}: $1$,
    \item \textit{max\_depth}: $5$,
    \item \textit{subsample}: $0.8658$,
    \item \textit{objective}: binary logistic.
\end{itemize}

\section{Ranges of inputs for surrogate training}
\label{sec:scaling}
Refer to \cite{anm6paper} for a description of the physical parameters that outline the subsequent ranges.
\begin{align*}
    a_{P_{g,t}} &\in [\underline{P}_g, \overline{P}_g]^{64} \quad \forall g \in \mathcal{D}_g\\
    a_{Q_{g,t}} &\in [\underline{Q}_g, \overline{Q}_g]^{64} \quad \forall g \in \mathcal{D}_g\\
    a_{P_{\text{DES},t}} &\in [\underline{P}_\text{DES}, \overline{P}_\text{DES}]^{64}\\
    a_{Q_{\text{DES},t}} &\in [\underline{Q}_\text{DES}, \overline{Q}_\text{DES}]^{64}\\
    P_{g,t}^{\text{(max)}} &\in [\underline{P}_g, \overline{P}_g]^{64} \quad \forall g \in \mathcal{D}_g\\
    P_{i}^{\text{(bus)}} &\in [\underline{P}_i^{\text{(bus)}}, \overline{P}_i^{\text{(bus)}}]^{64} \quad i = 1, \dots, 6\\
    Q_{i}^{\text{(bus)}} &\in [\underline{Q}_i^{\text{(bus)}}, \overline{Q}_i^{\text{(bus)}}]^{64} \quad i = 2, \dots, 6\\
    \text{SoC} &\in [\underline{\text{SoC}}, \overline{\text{SoC}}]^{64} \quad i = 2, \dots, 6
\end{align*}

\bibliographystyle{elsarticle-num}

\bibliography{example}






\end{document}